\documentclass[10pt,twocolumn,letterpaper]{article}

\usepackage{cvpr}              

\usepackage{graphicx}
\usepackage{amsmath}
\usepackage{amssymb}
\usepackage{booktabs}
\usepackage{multirow}
\usepackage{algorithm}
\usepackage{algorithmic}
\usepackage{array} 
\usepackage{xcolor}

\definecolor{cvprblue}{rgb}{0.21,0.49,0.74}
\usepackage[pagebackref,breaklinks,colorlinks,allcolors=cvprblue]{hyperref}

\def\paperID{*****} 
\def\confName{CVPR}
\def\confYear{2026}

\title{CBDiff:Conditional Bernoulli Diffusion Models for Image Forgery Localization}

\author{Zhou Lei\\
\and
Pan Gang\\
\and
Wang Jiahao
\and 
Sun Di
}

\begin{document}
\maketitle
\begin{abstract}
Image Forgery Localization (IFL) is a crucial task in image forensics, aimed at accurately identifying manipulated or tampered regions within an image at the pixel level. Existing methods typically generate a single deterministic localization map, which often lacks the precision and reliability required for high-stakes applications such as forensic analysis and security surveillance. To enhance the credibility of predictions and mitigate the risk of errors, we introduce an advanced Conditional Bernoulli Diffusion Model (CBDiff). 
Given a forged image, CBDiff generates multiple diverse and plausible localization maps, thereby offering a richer and more comprehensive representation of the forgery distribution. This approach addresses the uncertainty and variability inherent in tampered regions. Furthermore, CBDiff innovatively incorporates Bernoulli noise into the diffusion process to more faithfully reflect the inherent binary and sparse properties of forgery masks.
Additionally, CBDiff introduces a Time-Step Cross-Attention (TSCAttention), which is specifically designed to leverage semantic feature guidance with temporal steps to improve manipulation detection.
Extensive experiments on eight publicly benchmark datasets demonstrate that CBDiff significantly outperforms existing state-of-the-art methods, highlighting its strong potential for real-world deployment.
\end{abstract}    
\section{Introduction}
\label{sec:intro}

\begin{figure}[!h!t]
\begin{center}
\includegraphics[width=0.47\textwidth]{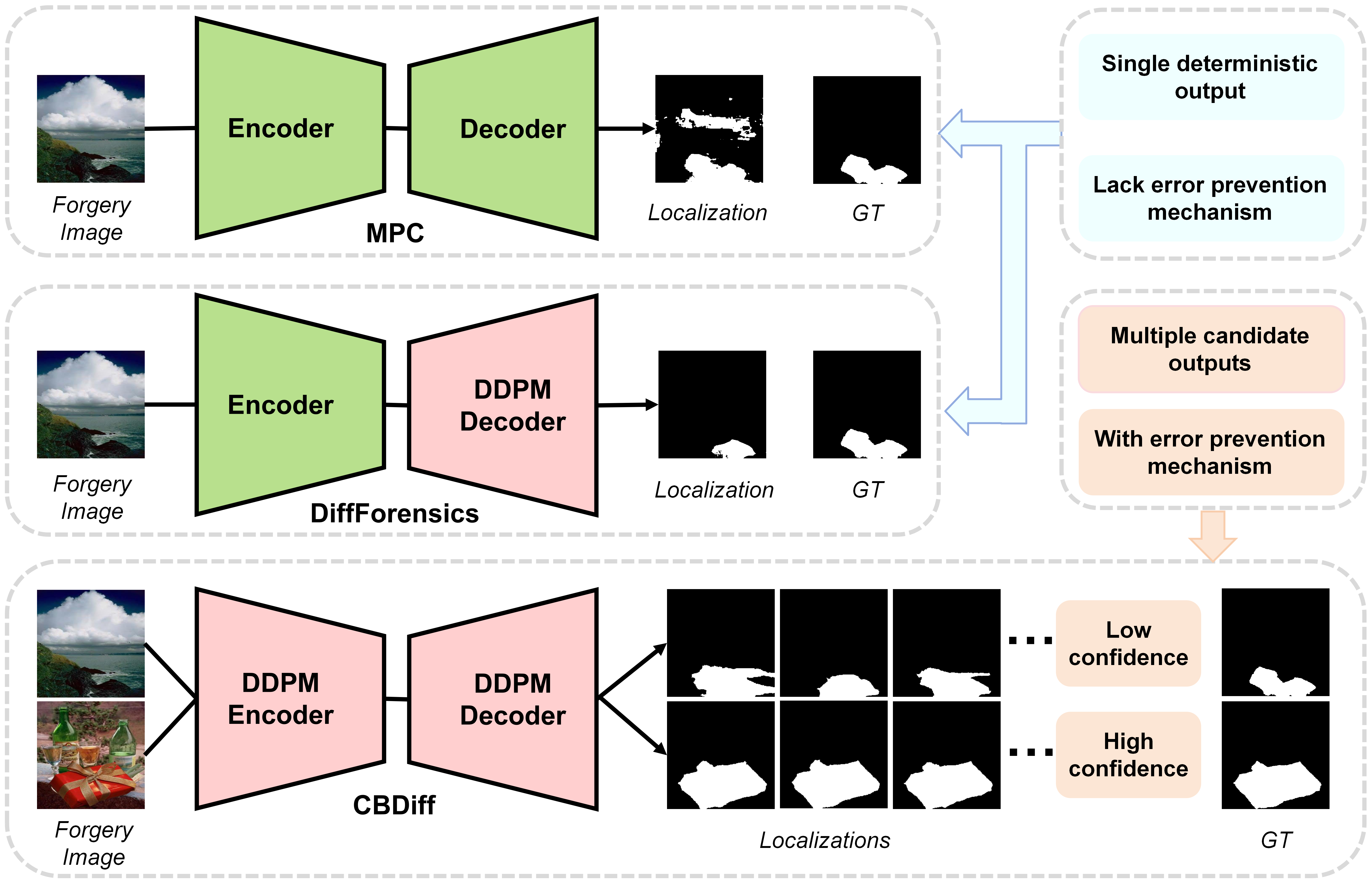}
\end{center}
\vspace{-3mm}
\caption{Comparison between the proposed method (CBDiff) and the previous approaches (MPC~\cite{MPC} and DiffForensics~\cite{DiffForensics}) in terms of forgery mask generation. Existing approaches output a single deterministic mask, limiting its robustness to uncertain cases. In contrast, CBDiff produces multiple candidate masks, capturing uncertainty and enhancing interpretability. For simple forgeries, CBDiff outputs are almost consistent; for complex cases, its predictions show greater diversity and coverage.}
\label{shoutu}
\end{figure}

With the rapid advancement of sophisticated image editing techniques, the ability to reliably detect and localize digital image manipulations has become increasingly vital. Image Forgery Localization (IFL) which aims to precisely identify tampered regions at the pixel level is a cornerstone of modern image forensics. Earlier IFL methods primarily relied on hand-crafted features and frequency-domain analysis techniques, such as Error Level Analysis (ELA)\cite{ELA}, Noise Inconsistency (NOI)\cite{NOI}, and Color Filter Array (CFA) pattern analysis~\cite{CFA}. Although effective under controlled conditions, these methods often face challenges in terms of robustness and generalization particularly when applied to images with complex backgrounds, high compression rates, or multiple manipulations—scenarios frequently encountered in real-world forensic applications. 

In recent years, the emergence of deep learning particularly Convolutional Neural Networks (CNNs) and Vision Transformers (ViTs) has substantially improved both the accuracy and automation of IFL tasks~\cite{ManTra-Net,RGB-N,MVSS-Net++,PSCC-Net,ObjectFormer,TBFormer,UP-Net,MPC,DiffForensics,CFA}. For example, ManTra-Net~\cite{ManTra-Net} employs an end-to-end encoder–decoder architecture for automatic forgery trace extraction, while RGB-N~\cite{RGB-N} enhances tampering discrimination by incorporating noise residuals alongside RGB channels. Advanced models such as MVSS-Net++\cite{MVSS-Net++} and PSCC-Net\cite{PSCC-Net} utilize multi-scale feature fusion to better adapt to varying manipulation scales. More recently, Transformer-based architectures like ObjectFormer~\cite{ObjectFormer}, TBFormer~\cite{TBFormer}, and UP-Net~\cite{UP-Net} have leveraged self-attention mechanisms to capture long-range dependencies, improving tamper localization in semantically complex scenes. However, most advancements have focused on detection accuracy and deterministic predictions, yet achieving absolute precision remains difficult in real-world scenarios characterized by uncertainty and variability in manipulated regions. 
Furthermore, there is a lack of explicit mechanisms for identifying prediction errors, which poses a significant risk in practical deployments.

Most existing methods are deterministic and generate only a single localization mask, as illustrated by MPC and DiffForensics in Fig.\ref{shoutu}. Such approaches fail to capture the inherent uncertainty and variability of tampered regions and lack mechanisms to prevent erroneous predictions. These limitations become particularly critical in high-stakes domains such as forensic investigations and media authenticity verification, where interpretability and reliability are indispensable. Recently, DiffForensics\cite{DiffForensics} leveraged diffusion models to simulate manipulation trajectories, but it remains essentially a feature regression model which adopts only the diffusion model decoder with a fixed time step. Consequently, it does not explicitly produce diverse segmentation outcomes or adequately represent spatial uncertainty.

To address these limitations, we propose CBDiff, a novel Conditional Bernoulli Diffusion Model specifically designed for image forgery localization. Compared with existing methods, CBDiff generates multiple localization results for the same forged image, as illustrated by CBDiff in Fig.~\ref{shoutu}. High consistency among these results indicates a high level of confidence in the model’s output. Conversely, significant variations suggest lower reliability, serving as a cautionary signal to avoid potential errors in referencing, and thus minimizing the risk of using incorrect results. This diversity enhances robustness and interpretability while also facilitating human verification and confidence assessment. By replacing conventional Gaussian noise with Bernoulli noise, CBDiff better models the binary and sparse nature of tampering masks, simplifying the learning process. Furthermore, we propose a Time-Step Cross-Attention (TSCAttention) mechanism that incorporates multi-scale semantic features extracted from a DINO-pretrained backbone into the diffusion model as conditional guidance. This design improves the effectiveness of the model in semantically complex scenarios. Additionally, the progressive generation process of diffusion models allows CBDiff to maintain strong performance even with low-resolution or subtly manipulated images. In summary, our main contributions are as follows:
\begin{itemize}
    \item We introduce a new task: generating multiple plausible localizations of tampered regions within a single forged image. To tackle this task, we propose CBDiff, a custom-designed method that captures the uncertainty and variability of forged images and incorporates an error prevention mechanism.
    \item We innovatively incorporate Bernoulli noise into the diffusion process to more faithfully reflect the inherent binary and sparse properties of forgery masks.
    \item We design a Time-Step Cross-Attention (TSCAttention) module to condition the diffusion process on rich semantic features extracted via a DINO-pretrained backbone.
\end{itemize}

\section{Related Work}
\noindent \textbf{Image Forgery Localization.} Image forgery localization has long been a core task in digital forensics, aiming to identify manipulated regions with pixel-level precision. Early approaches predominantly relied on hand-crafted low-level features, such as Color Filter Array (CFA) inconsistencies~\cite{CFA}, Error Level Analysis (ELA)~\cite{ELA}, noise residuals (NOI)~\cite{NOI}, JPEG compression artifacts~\cite{CAT-Net}, and resampling artifacts~\cite{popescu2004statistical}. These methods were effective for specific manipulation types (e.g., splicing or copy-move) and performed reasonably well under ideal conditions (e.g., low compression and minimal post-processing). However, their limited robustness and generalization ability often led to performance degradation when faced with complex image content, diverse tampering strategies, and heavy post-processing, which restricts their applicability in real-world forensic scenarios.

With the advent of deep learning, forgery localization methods have increasingly shifted towards deep neural network-based architectures, typically formulating the problem as a pixel-wise binary segmentation task. Representative works such as ManTra-Net~\cite{ManTra-Net} employ an end-to-end convolutional encoder-decoder framework to automatically extract manipulation traces, while RGB-N~\cite{RGB-N} enhances the detection of high-frequency artifacts by combining residual noise maps with RGB inputs. More advanced multi-scale architectures, including MVSS-Net++~\cite{MVSS-Net++}, PSCC-Net~\cite{PSCC-Net} and SSR-IFL~\cite{sheng2025exploring}, integrate semantic and frequency-domain features, substantially improving localization accuracy.

The introduction of Vision Transformers has further boosted performance by capturing long-range dependencies through self-attention mechanisms. Notable examples include ObjectFormer~\cite{ObjectFormer}, TBFormer~\cite{TBFormer}, UP-Net~\cite{UP-Net}, TruFor~\cite{Trufor} and SAFIRE~\cite{Safire}, which are better equipped to model complex backgrounds and subtle manipulations. Nevertheless, these approaches remain fundamentally discriminative models that output a single deterministic mask, making it difficult to capture the inherent uncertainty and diversity of tampered regions. This limitation can be particularly problematic in forensic applications, where interpretability and confidence estimation are crucial. Recently, DiffForensics~\cite{DiffForensics} was the first to introduce diffusion models into the forgery localization domain. However, its design remains focused on boundary refinement and feature recovery, without explicitly modeling the distributional uncertainty of forgery masks or generating diverse candidate predictions.

\noindent \textbf{Denoising Diffusion Probabilistic Models}. Denoising Diffusion Probabilistic Models(DDPMs)~\cite{ho2020denoising} have recently demonstrated remarkable success in image generation and modeling tasks. The core principle is to model complex data distributions through a forward diffusion process, which gradually adds noise to the data, and a parameterized reverse denoising process that reconstructs the data. Compared with GANs, DDPMs are more stable to train and often produce higher-quality images. Diffusion models have been widely applied in tasks such as image generation~\cite{ho2020denoising, dhariwal2021diffusion, ramesh2022hierarchical, song2020denoising}, inpainting~\cite{chung2022come, rombach2022high}, and editing~\cite{avrahami2022blended, choi2021ilvr}, and have recently been extended to discriminative tasks such as image segmentation~\cite{baranchuk2021label, brempong2022denoising} and anomaly detection~\cite{wolleb2022diffusion, wyatt2022anoddpm}. However, most of these works employ diffusion models either for data augmentation or as feature extractors. Research has shown that diffusion models are inherently advantageous in modeling data diversity and multi-scale details; their iterative generative mechanism allows them to effectively capture both the underlying structure and fine-grained features of images.
\section{Methods}

\begin{figure*}[!h!t]
\begin{center}
\includegraphics[width=0.95\textwidth]{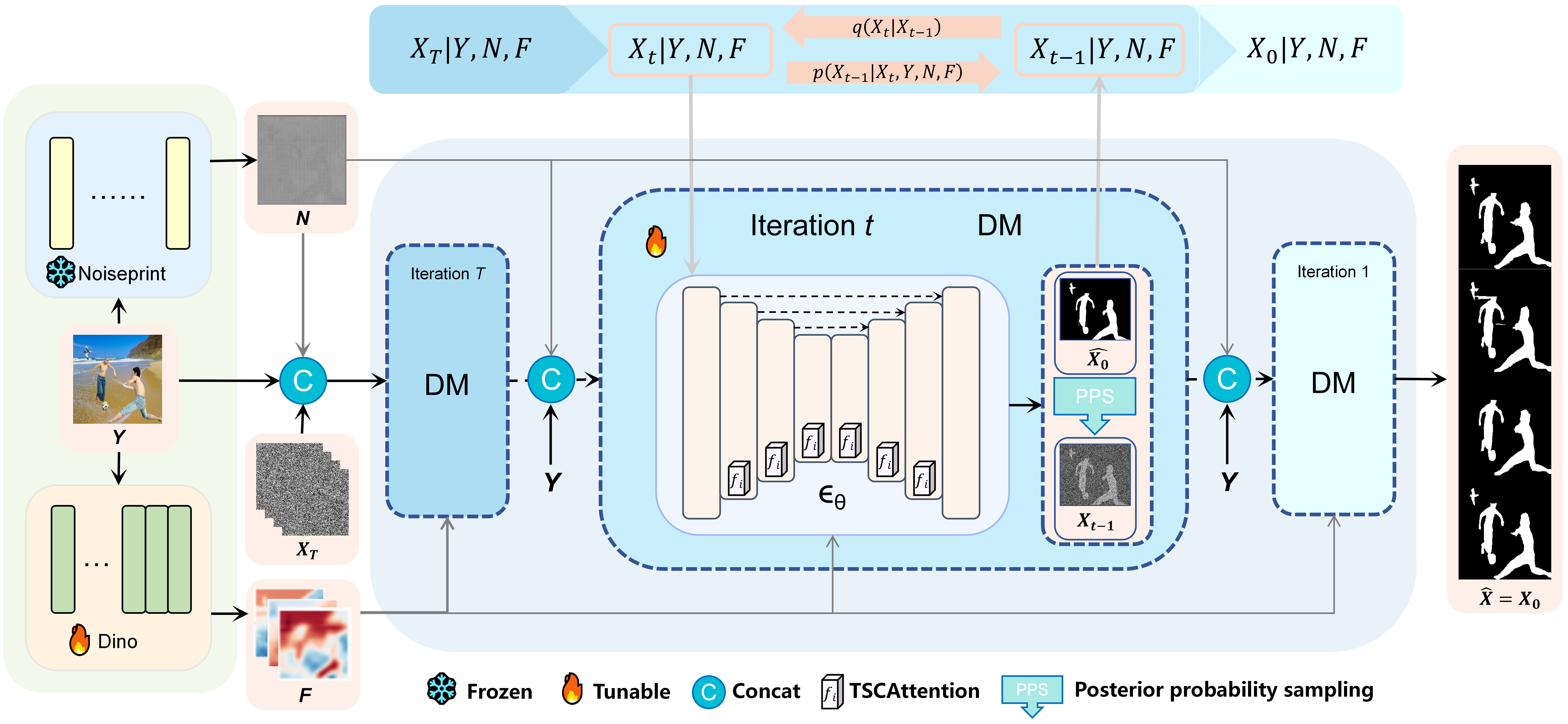}
\end{center}
\vspace{-3mm}
\caption{Overall architecture of CBDiff framework. Given an input image $Y$, semantic features $F$ are extracted via a DINO-pretrained backbone~\cite{caron2021emerging}, while noise residuals $N$ are computed using a Noiseprint++ extractor~\cite{Trufor}. These are fused as conditional inputs to guide the denoising process of the Bernoulli-based diffusion model (DM). Starting from pure Bernoulli noise $X_T$, the model progressively refines predictions across $T$ iterations to generate diverse candidate masks $\hat{X}_0$. The TSAttention module injects multi-scale semantic features $f_i$ into the diffusion process, enhancing forgery localization accuracy and robustness.}
\label{zhutu}
\end{figure*}

\subsection{Background and Definitions}
DDPMs~\cite{ho2020denoising} are latent variable models defined as $p_{\theta}(x_{0})=\int p_{\theta(x0:T)dx_{1:T}}$, which consist of a forward diffusion process and a reverse denoising process. In the forward process, DDPM defines a fixed, non-learnable Markov chain $q(x_{1:T}|x_{0})=\prod_{t=1}^{T}q(x_{t}|x_{t-1})$, that gradually perturbs the input data. In the reverse process, a parameterized model is trained to map noisy latent variables back to the data distribution $p_{\theta}(x_{0:T})=p(x_{T})\prod_{t=1}^T p_{\theta}(x_{t-1}|x_{t})$, where $x_0 \in \mathbb{R}^D$ is the observed variable and ${x}_{t=1}^T$ are latent variables of the same dimensionality. The original DDPM assumes that the transition distributions in both forward and reverse processes are Gaussian with diagonal covariance and that $p(x_T)$ follows a standard multivariate normal distribution. However, these assumptions become inadequate when $x_0$ consists of elements from discrete, unordered sets, such as binary forgery localization masks. Modeling such discrete data using Gaussian noise leads to distributional mismatches and degraded generation quality.  

\subsection{Bernoulli Diffusion Model}
To better model binary forgery localization masks, we propose a Bernoulli-based diffusion process. A localization mask can be represented as $X_{0} \in \{0,1\}^{H \times W \times 2}$, where $H \times W$ denotes the image resolution and the last dimension encodes the foreground–background label. This structure naturally corresponds to multiple dependent Bernoulli variables. Therefore, using Bernoulli noise rather than Gaussian noise is more suitable for modeling the discrete mask distribution, which directly addresses the limitation of standard DDPMs. Inspired by~\cite{hoogeboom2021argmax}, all latent variables $X_{1:T}$ and the transition distributions of the forward and reverse processes are parameterized with Bernoulli distributions. The forward process corrupts the mask $X_{t-1}$ element-wise to obtain the parameters of the distribution for $X_t$:
\begin{equation}
q(X_{t}|X_{t-1})=\mathcal{B}(X_{t};\frac{\beta_{t}}{2}\mathbf{1}+(1-\beta_{t})X_{t-1}),
\end{equation}
where $\mathbf{1}$ is an all-ones tensor of size ${H \times W \times 2}$ and $\alpha_t = 1 - \beta_t \in (0,1)$ represents the probability of preserving the label at step $t$. Because this process is a Markov chain, any $X_t$ can be expressed in closed form given the clean mask $X_0$:
\begin{equation}
\begin{aligned}
q(X_{t}|X_{0})&=\mathcal{B}(X_{t};\frac{1-\overline{\alpha}_{t}}{2}\mathbf{1}+\overline{\alpha}_{t}X_{0}), \\
\bar{\alpha}_t&=\prod_{i=1}^t \alpha_i.
\end{aligned}
\end{equation}

The posterior distribution of the previous state, required for reverse sampling, can be derived using Bayes’ theorem:
\begin{equation}
\begin{aligned}
q(X_{t-1}|X_{t},X_{0})&=\mathcal{B}(X_{t-1};\mathbf{\theta}_{post}(X_{t},X_{0})), \\
\mathbf{\theta}_{post}(X_{t},X_{0})&=\frac{\tilde{\mathbf{\theta}}(X_{t},X_{0})}{\sum^2_{k=1}\tilde{\theta}(X_{t},X_{0})[k]},\\
\tilde{\mathbf{\theta}}(X_{t},X_{0})&=[\frac{1-\overline{\alpha}_{t}}{2}\mathbf{1}+\overline{\alpha}_{t}X_{0}]\odot[\frac{1-\alpha_{t}}{2}\mathbf{1}+\alpha_{t}X_{t}].
\end{aligned}
\end{equation}
The reverse process also follows an element-wise Bernoulli distribution:
\begin{equation}
p_{\theta}(X_{t-1}|X_{t})=\mathcal{B}(X_{t-1};\hat{\mathbf{P}}_{t-1}),
\end{equation}
Since the network must be differentiable, we cannot directly predict $\hat{X}_{t-1}$; instead, we follow Ho et al. [19] and predict the clean mask distribution $\hat{\mathbf{P}}_0 \in [0,1]^{H \times W \times 2}$ at each step:
\begin{equation}
\begin{aligned}
\hat{\mathbf{P}}_{t-1}&=\sum_{k=1}^{2}\mathbf{\theta}_{post}(X_{t},\overline{X}_{0}[k])*\hat{\mathbf{P}}_{0}[k],\\
\hat{\mathbf{P}}_{0}&=\phi_{\theta}(X_{t},t),
\end{aligned}
\end{equation}
where $\bar{X}_0$ denotes all possible states. This formulation maintains a consistent output space across timesteps, which improves training stability and convergence. This Bernoulli diffusion addresses the gap left by Gaussian-based DDPMs by aligning the noise model with the discrete nature of forgery masks, thereby reducing information loss in the forward process and improving reconstruction quality in the reverse process.

\subsection{Conditional Bernoulli Diffusion Model}
Forgery localization can be formulated as modeling the conditional distribution $q(X_0 \mid Y)$, where $Y \in \mathbb{R}^{H \times W \times 3}$ is the forged input image and $X_0 \in {0,1}^{H \times W}$ is the corresponding mask. Directly learning this conditional distribution using traditional discriminative models is challenging because forged regions exhibit large diversity and high subtlety. To tackle this, we design the Conditional Bernoulli Diffusion Model (CBDiff), which decouples the complex interactions between image pixels and masks through a diffusion-based generative modeling framework. As shown in Fig.~\ref{zhutu}, we incorporate three complementary sources of information as conditional inputs: The forged image $Y$ itself. Noiseprint++~\cite{Trufor} features $N \in \mathbb{R}^{H \times W \times 1}$ that capture camera trace inconsistencies indicative of forgeries. Multi-scale semantic features $F = {f_i \in \mathbb{R}^{H_i \times W_i \times C_i}}$ extracted by a pretrained DINO transformer\cite{caron2021emerging}, which provide high-level contextual understanding. This conditioning is realized by injecting $(Y, N, F)$ into the denoising network $\phi_\theta(X_t, t, Y, N, F)$. $\phi_\theta$ is based on UNet framework~\cite{dhariwal2021diffusion}, but the self-attention layers in its deepest three blocks are replaced with our Cross Attention with Time Steps (TSCAttention) modules (Fig.~\ref{tsattention}).

The forward process remains the same as the standard Bernoulli diffusion, as each step depends only on the previous state:
\begin{equation}
q(X_{1:T} \mid X_0, Y, N, F) = q(X_{1:T} \mid X_0).
\label{eq:forward}
\end{equation}
In the reverse process, we condition every denoising step on the image and extracted features:
\begin{equation}
p_\theta(X_{0:T} \mid Y, N, F) = p(X_T) \prod_{t=1}^T p_\theta(X_{t-1} \mid X_t, Y, N, F).
\label{eq:reverse}
\end{equation}
At each step, the current mask state $X_t$, forged image $Y$, and noiseprint features $N$ are concatenated along the channel dimension as input to $\phi_\theta$, while $F$ is used in TSCAttention for cross-scale semantic enhancement.

\subsection{TSCAttention: Cross Attention with Time Steps}
\begin{figure}[t!b]
\centering
\includegraphics[width=3in]{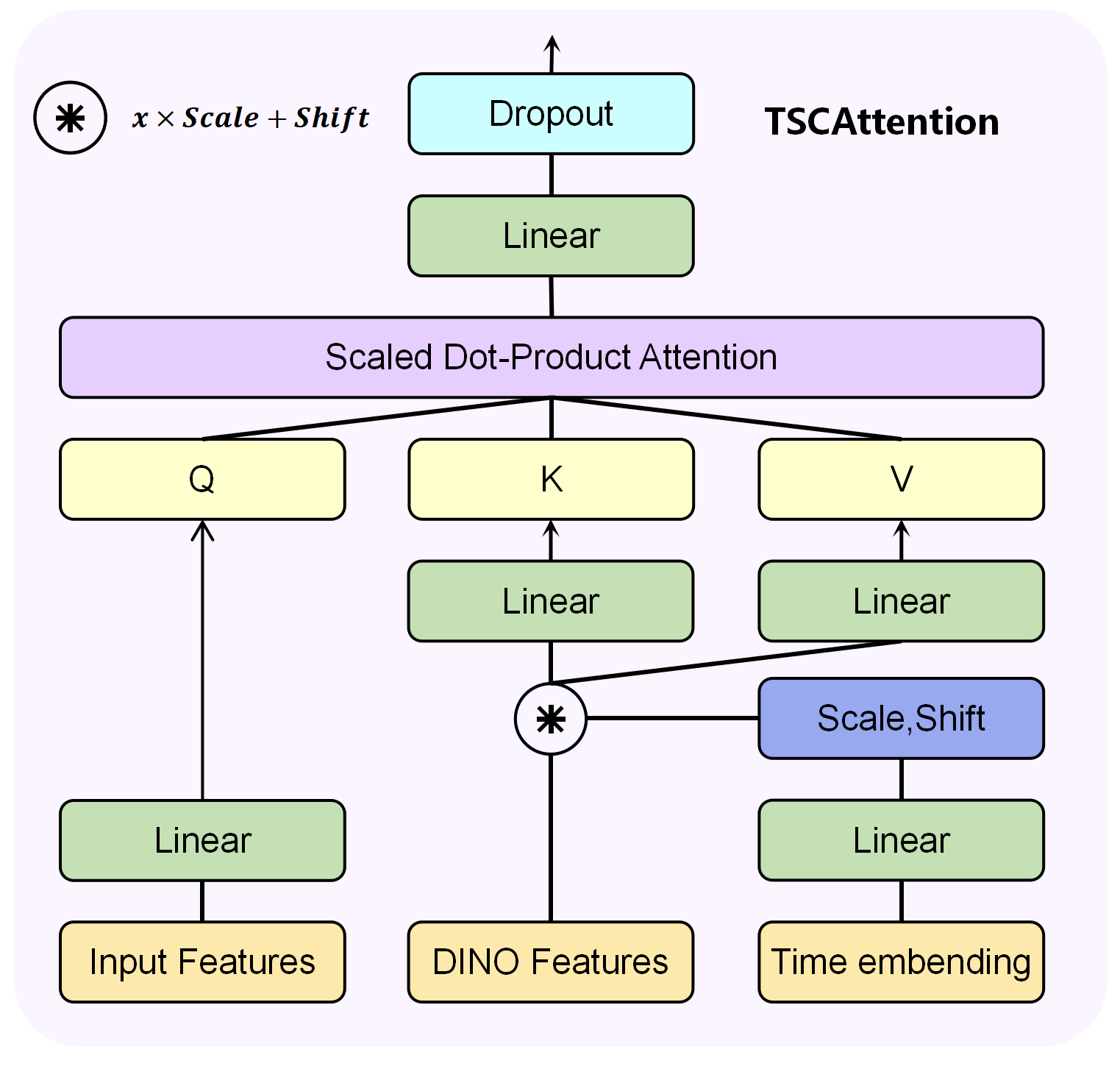}
\caption{Architecture of Time-Step Cross-Attention (TSCAttention) module. Input features are linearly projected to obtain the query $Q$, while modulated DINO features serve as keys $K$ and values $V$ after linear transformation. A learnable scale-and-shift operation, modulated by time embeddings, is applied to the DINO features. The resulting attention map is computed via scaled dot-product attention and passed through a linear projection and dropout layer. This design enables effective injection of semantic and temporal information into the diffusion process.}
\label{tsattention}
\end{figure}
Cross Attention (CA)~\cite{chen2021crossvit} provides an effective mechanism to fuse two embedding sequences of identical dimensionality, thereby facilitating richer feature interaction and representation enhancement. However, directly applying CA in our diffusion-based framework reveals a critical limitation: the visual backbone (e.g., DINO~\cite{caron2021emerging}) operates in a static manner and lacks explicit timestep information, which makes it difficult to dynamically adapt to the varying noise levels and semantic uncertainty across different diffusion stages. This limitation restricts the network’s ability to model the evolution of manipulated regions throughout the denoising process.

To overcome this issue, we propose Cross Attention with Time Steps (TSCAttention) as shown in Fig.\ref{tsattention}, an enhanced cross-attention module that explicitly incorporates timestep embeddings into the feature fusion process. Unlike conventional CA, TSCAttention enables semantic features to be modulated according to the temporal evolution of the diffusion process, allowing the network to adaptively adjust its focus at each denoising step. Specifically, the input features, derived from the current noisy mask state and auxiliary modalities (e.g., forged image and noise residuals), are first projected to form query vectors. Meanwhile, the timestep embedding is transformed into a pair of scale and shift parameters, which are applied to the DINO-derived multi-scale semantic features through element-wise affine modulation. This operation explicitly injects temporal sensitivity into the semantic representation, producing time-aware features that are subsequently mapped into key and value vectors. Together with the queries, these temporally modulated features are fed into the scaled dot-product attention mechanism, and the resulting output is further refined by a linear projection and dropout layer before being propagated through the reverse diffusion network. By introducing timestep-aware modulation into cross-attention, TSCAttention achieves a deep fusion of temporal and semantic cues. This design not only enables the attention mechanism to adapt its focus to the noise level at each diffusion stage but also enhances the modeling of dynamic semantic evolution, ultimately improving the localization accuracy and contextual robustness of forged region reconstruction under challenging conditions.

\subsection{Training and Inference}
The proposed diffusion model is optimized using variational inference. Given the ground-truth mask $X_0$, the posterior distribution $q(X_{t-1}|X_{t},X_{0})$ is tractable and easy to compute, which allows us to construct a low-variance evidence lower bound (ELBO) via the KL-divergence formulation with Eq.~\ref{eq:forward} and Eq.~\ref{eq:reverse}:
\begin{equation}
\begin{aligned}
\log p_{\theta}(&X_{0}|Y,N,F)\geq\mathbb{E}_{q}[\log p_{\theta}(X_{0}|X_{1},Y,N,F)\\
&-\sum_{t=2}^T KL(q(X_{t-1}|X_{t},X_{0})||p_{\theta}(X_{t-1}|X_{t},Y,N,F))\\
&-KL(q(X_{T}|X_{0})||p_{\theta}(X_{T}|Y,N,F))].
\end{aligned}
\end{equation}
The first two terms can be optimized via standard stochastic gradient descent. The third term, however, naturally vanishes in our Bernoulli diffusion design: because the forward process converges to a uniform Bernoulli distribution $\lim_{t\rightarrow+\infty}q(X_{t}|X_{0})=\mathcal{B}(X;0.5)$, we explicitly fix the terminal prior as $p(X_{T}|Y,N,F)=p(X_{T})=\mathcal{B}(X_{T};0.5)$. This ensures that $q(X_{t}|X_{0})\approx p(X_{t}|Y,N,F)$, making the third KL term in the ELBO approximately zero. Based on this property, the overall loss function can be written as: 
\begin{equation}
L = 
\begin{cases} 
KL(q(X_{t-1}|X_{t},x_{0})||p_{\theta}(X_{t-1}|X_{t},Y,N,F)), &\text{if }t\geq2 \\
-\sum _{k=1}^2X_{0}[k]\log \hat{X}_{0}[k], &\text{if }t=1
\end{cases},
\end{equation}
where $\hat{X}_0$ denotes the predicted posterior at the final denoising step. The detailed training and inference procedure is summarized in Alg.~\ref{alg:training} and Alg.~\ref{alg:inference}.

\begin{algorithm}[t]
\caption{Training a Conditional Bernoulli Diffusion Model (CBDM) with $T$ steps}
\label{alg:training}
\begin{algorithmic}[1]
\REQUIRE Forgery images $Y \in \mathbb{R}^{H \times W \times 3}$ and location label maps $X_0 \in \mathbb{R}^{H \times W \times 2}$ from the training dataset
\REPEAT
    \STATE Sample $t \sim \mathrm{Uniform}(\{1, \dots, T\})$
    \STATE Sample $\mathbf{X}_t \sim q(\mathbf{X}_t \mid \mathbf{X}_0, t)$
    \STATE Compute $\hat{\mathbf{P}}_0 \leftarrow \phi_\theta(\mathbf{X}_t, Y, N, F, t)$
    \IF{$t > 1$}
        \STATE $\hat{\mathbf{P}}_{t-1} = \sum_{k=1}^{2} \mathbf{\theta}_{post}(\mathbf{X}_t, \overline{\mathbf{X}}_0[k]) \cdot \hat{\mathbf{P}}_0[k]$
        \STATE $L \leftarrow KL\big(q(\mathbf{X}_{t-1} \mid \mathbf{X}_t, \mathbf{X}_0) \parallel p_\theta(\mathbf{X}_{t-1} \mid \hat{\mathbf{P}}_{t-1})\big)$
    \ELSE
        \STATE $\hat{\mathbf{X}}_0 \sim \mathcal{B}(\hat{\mathbf{X}}_0; \hat{\mathbf{P}}_0)$
        \STATE $L \leftarrow -\sum_{k=1}^{2} \mathbf{X}_0[k] \log \hat{\mathbf{X}}_0[k]$
    \ENDIF
    \STATE $\theta \leftarrow \theta - \nabla_\theta L$ \COMMENT{Gradient descent}
\UNTIL{converged}
\end{algorithmic}
\end{algorithm}

\begin{algorithm}[t]
\caption{Inference from a Conditional Bernoulli Diffusion Model (CBDM) with $T$ steps}
\label{alg:inference}
\begin{algorithmic}[1]
\REQUIRE Forgery images $Y \in \mathbb{R}^{H \times W \times 3}$ and trained network $\phi_\theta$ from Algorithm~\ref{alg:training}
\STATE Initialize $\mathbf{X}_T \sim \mathcal{B}(\mathbf{X}_T; 0.5)$ \COMMENT{Sampling prior}
\STATE Set $\mathbf{X}_t \leftarrow \mathbf{X}_T$
\FOR{$t = T, T-1, \dots, 1$}
    \STATE $\hat{\mathbf{P}}_0 \leftarrow \phi_\theta(\mathbf{X}_t, Y, N, F, t)$
    \IF{$t > 1$}
        \STATE $\hat{\mathbf{P}}_{t-1} = \sum_{k=1}^{2} \mathbf{\theta}_{post}(\mathbf{X}_t, \overline{\mathbf{X}}_0[k]) \cdot \hat{\mathbf{P}}_0[k]$
        \STATE $\mathbf{X}_{t-1} \sim \mathcal{B}(\mathbf{X}_{t-1} \mid \hat{\mathbf{P}}_{t-1})$ \COMMENT{Sampling}
    \ELSE
        \STATE $\mathbf{X}_t \leftarrow \arg\max_{k=1}^{2} \hat{\mathbf{P}}_0[:, k]$ \COMMENT{Final prediction}
    \ENDIF
\ENDFOR
\RETURN $\mathbf{X}_t$
\end{algorithmic}
\end{algorithm}
\section{Experiments}
\subsection{Experimental Setup}

\noindent\textbf{Dataset.}
To ensure a fair and comprehensive comparison with existing methods, we follow the experimental protocol outlined in~\cite{MPC}, and construct two distinct training configurations. Although some baselines do not release source code, we replicate their data settings and evaluation protocols as closely as possible. The details are as follows:

\begin{itemize}
    \item \textbf{Protocol 1:} As most existing approaches are trained on datasets ranging from 50K to 100K images, we pretrain our model on 60,000 forged images randomly sampled from the CAT-Net dataset. We then fine-tune all models using a 3:1 training-to-testing split on CASIAv1, NIST, Coverage, and IMD20, and perform unified evaluation on the test sets.
    
    \item \textbf{Protocol 2:} We strictly follow the training configuration used by MPC, evaluating on eight public image forgery datasets. Due to computational constraints, comparisons under this protocol are limited to CAT-Net, TruFor, and MPC, which share the same training setup.
\end{itemize}

\noindent\textbf{Metrics.}
Forgery localization accuracy is evaluated using the F1-score and area under the ROC curve (AUC). A binary mask is generated from the predicted probability map using a threshold of 0.5. To address the variation in image counts across datasets, we report weighted average performance across datasets:
\begin{equation}
\text{Ave} = \frac{\sum_{i=1}^{N} \text{Metric}_i \times \text{Num}_i}{\sum_{i=1}^{N} \text{Num}_i},
\end{equation}
where $\text{Metric}_i$ denotes the average score (F1 or AUC) on the $i$-th dataset, and $\text{Num}_i$ represents the number of images in that dataset. This metric reflects performance at the image level and avoids bias introduced by treating datasets equally regardless of size.

\noindent\textbf{Training Details.}
We train our models on eight NVIDIA A800 GPUs (80GB each). The initial learning rate is set to $1 \times 10^{-4}$ and decayed polynomially to a minimum of $1 \times 10^{-6}$. We adopt the AdamW optimizer with $\beta_1 = 0.9$, $\beta_2 = 0.999$, a weight decay of 0.01, and a batch size of 64.

Models are trained for 500 epochs under Protocol 1, and for 50 epochs under Protocol 2. Data augmentation strategies follow those used in MPC, including random cropping, horizontal flipping, and color jittering.

\begin{figure}
    \centering
    \setlength{\tabcolsep}{1pt}
    \renewcommand{\arraystretch}{0.5}
    \resizebox{0.47\textwidth}{!}{
        \begin{tabular}{ccccccccc}
            \raisebox{2\height}{\rotatebox[origin=c]{90}{Image}} &
            \includegraphics[width=0.12\textwidth]{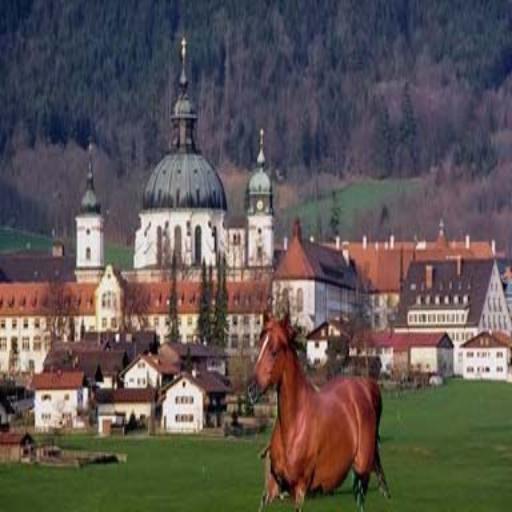} & \includegraphics[width=0.12\textwidth]{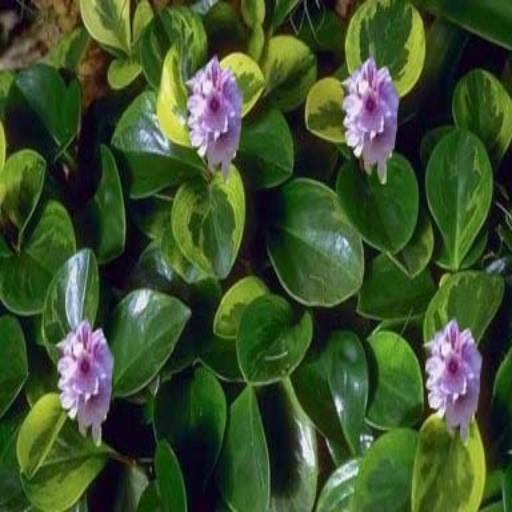} &
            \includegraphics[width=0.12\textwidth]{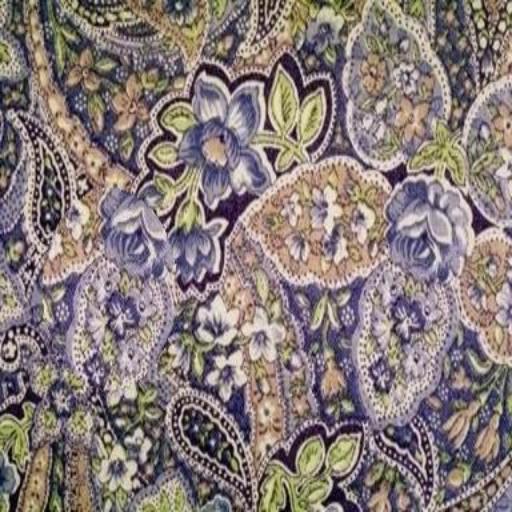} & \includegraphics[width=0.12\textwidth]{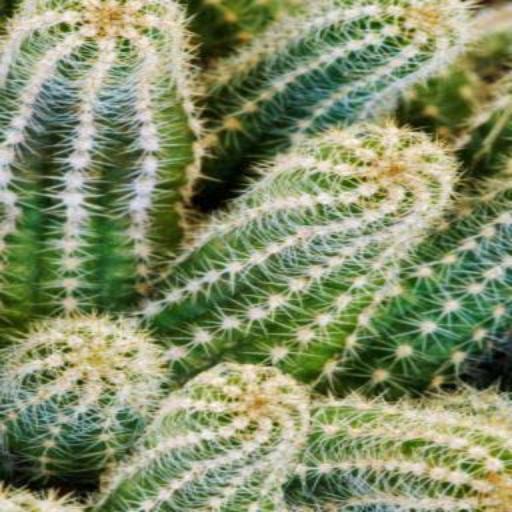} &
            \includegraphics[width=0.12\textwidth]{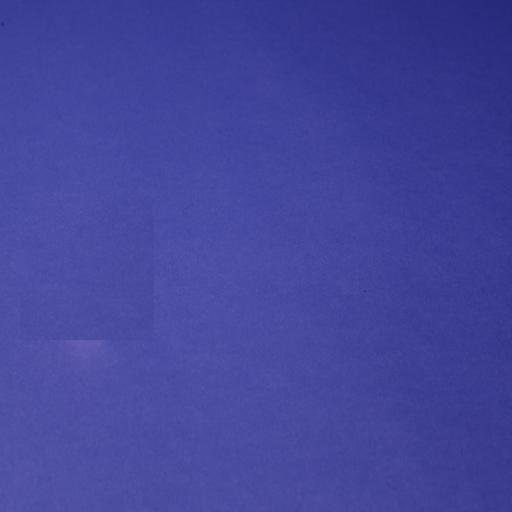} & \includegraphics[width=0.12\textwidth]{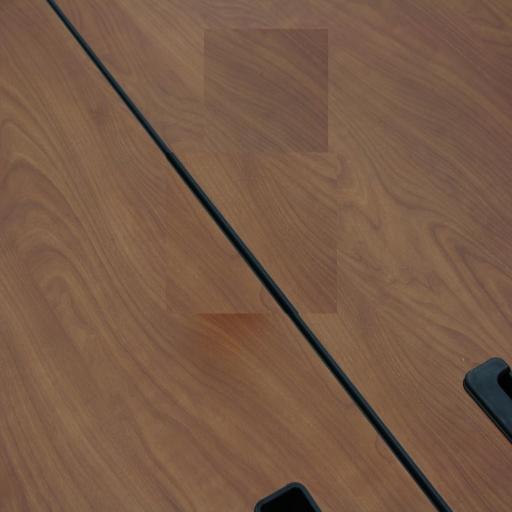} & \includegraphics[width=0.12\textwidth]{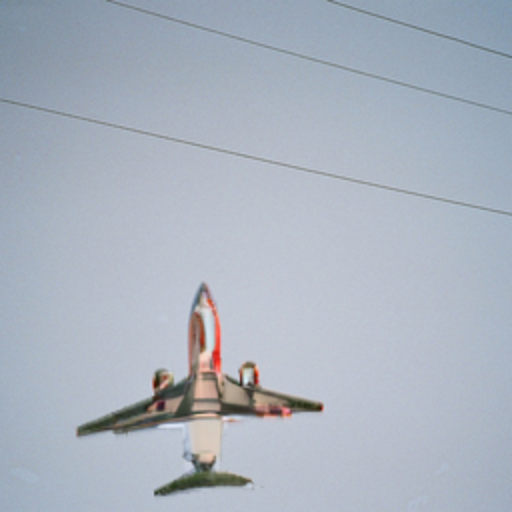} & \includegraphics[width=0.12\textwidth]{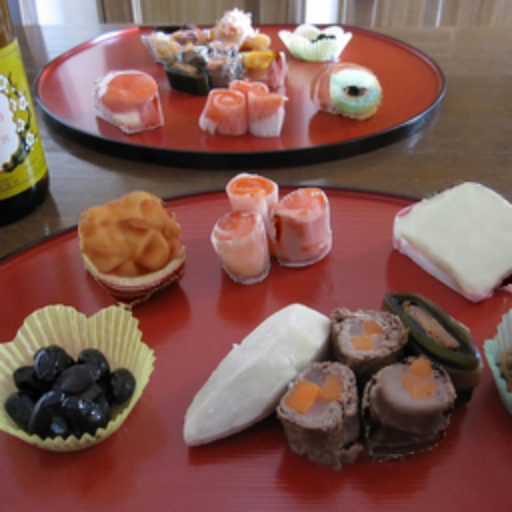} \\
    
            \raisebox{2.6\height}{\rotatebox[origin=c]{90}{GT}} &
            \includegraphics[width=0.12\textwidth]{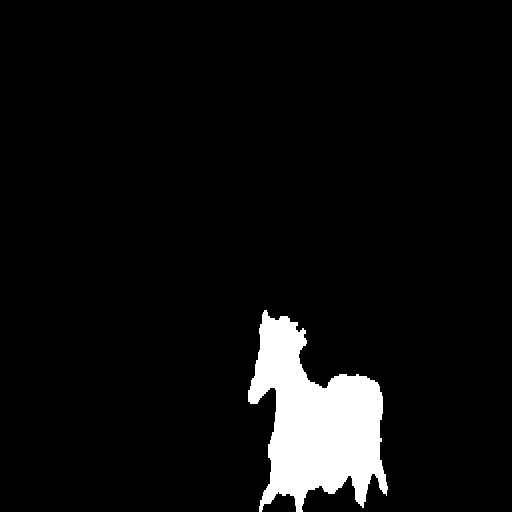} & \includegraphics[width=0.12\textwidth]{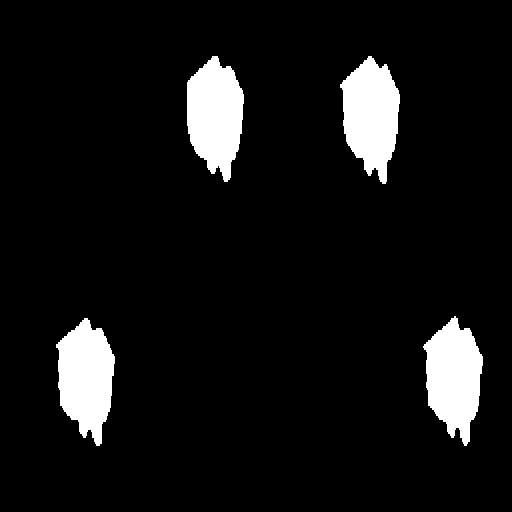} &
            \includegraphics[width=0.12\textwidth]{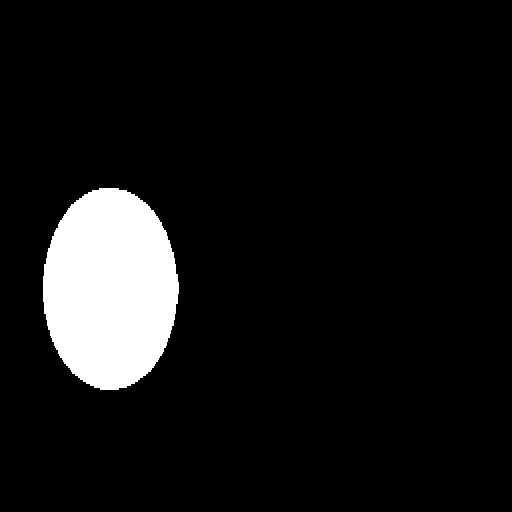} & \includegraphics[width=0.12\textwidth]{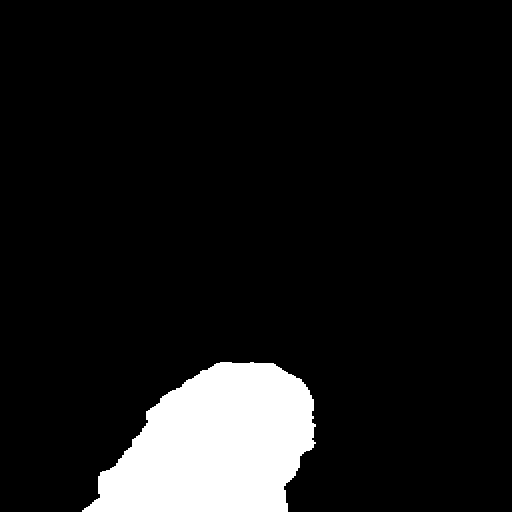} & \includegraphics[width=0.12\textwidth]{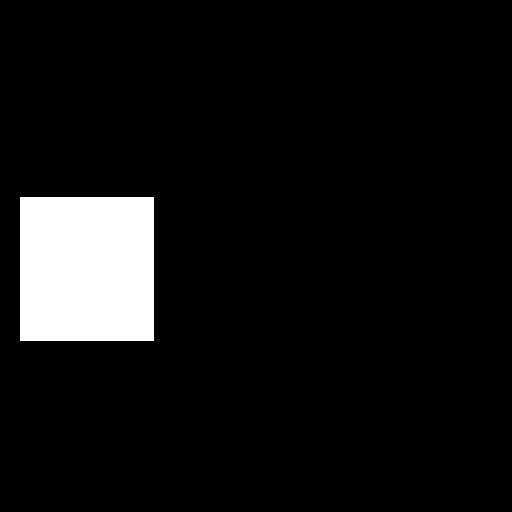} & 
            \includegraphics[width=0.12\textwidth]{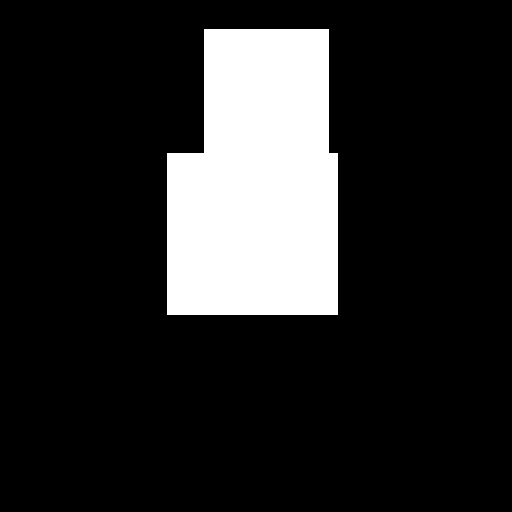} &
            \includegraphics[width=0.12\textwidth]{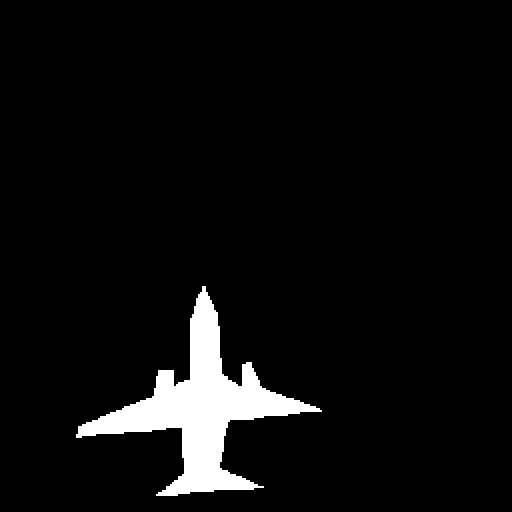} & \includegraphics[width=0.12\textwidth]{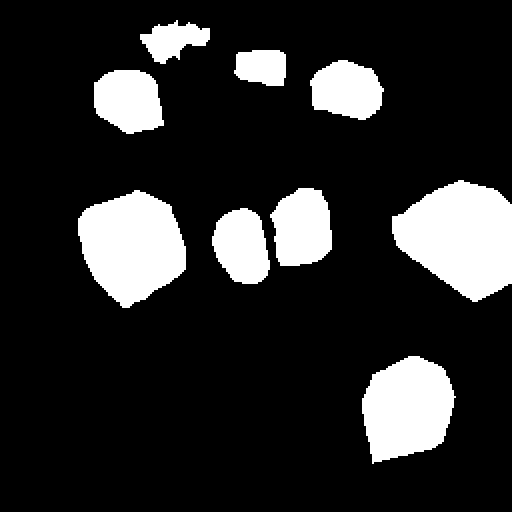} \\
            
            \raisebox{1.4\height}{\rotatebox[origin=c]{90}{CAT-Net}} &
            \includegraphics[width=0.12\textwidth]{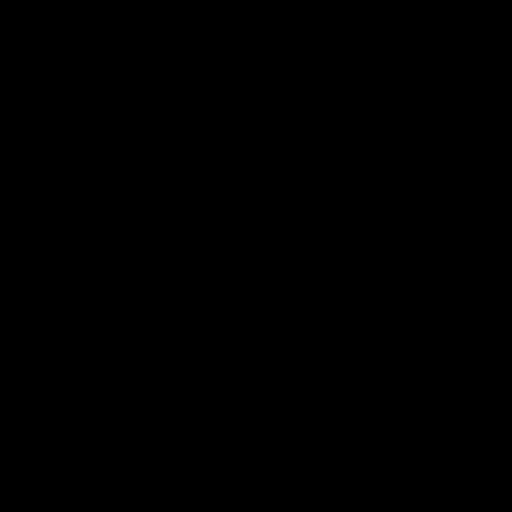} & \includegraphics[width=0.12\textwidth]{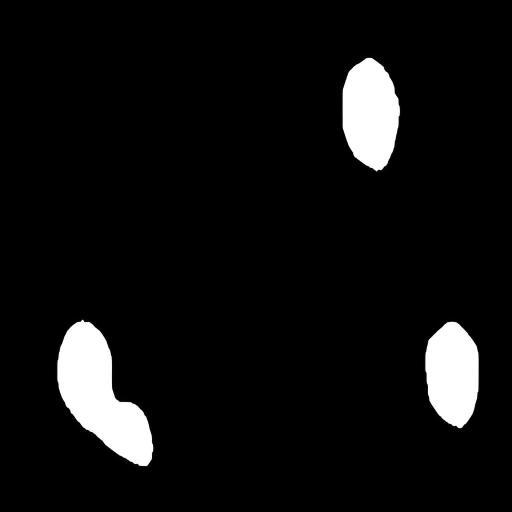} &
            \includegraphics[width=0.12\textwidth]{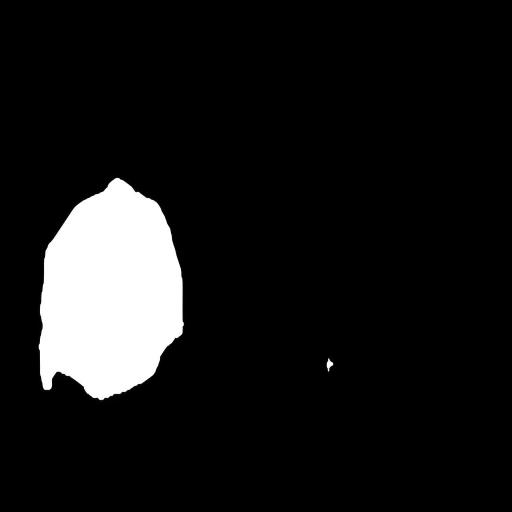} & \includegraphics[width=0.12\textwidth]{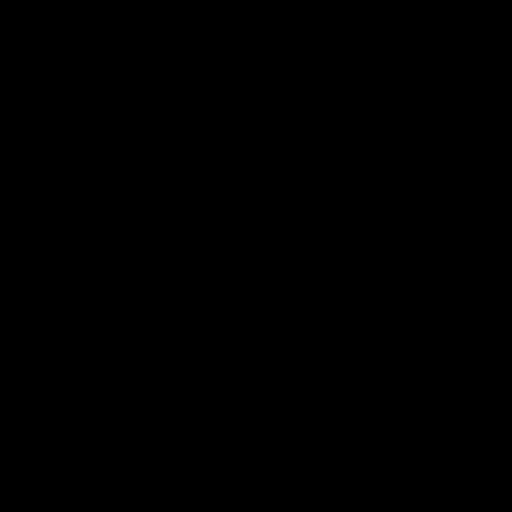} & \includegraphics[width=0.12\textwidth]{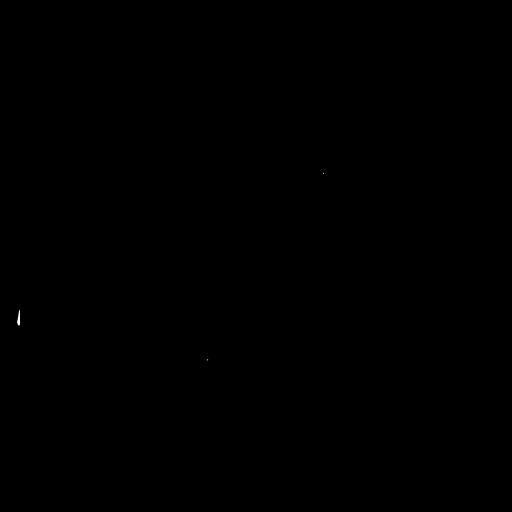} &
            \includegraphics[width=0.12\textwidth]{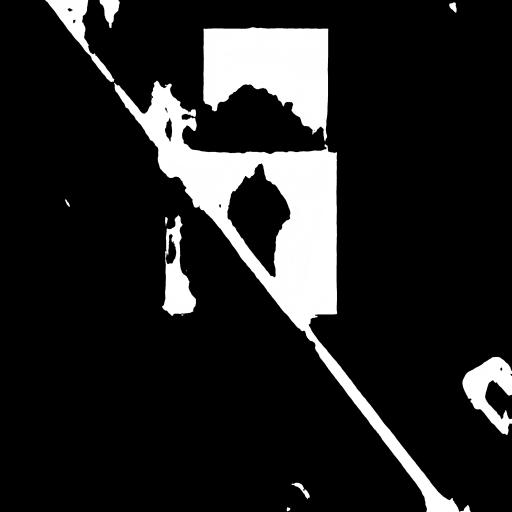} &
            \includegraphics[width=0.12\textwidth]{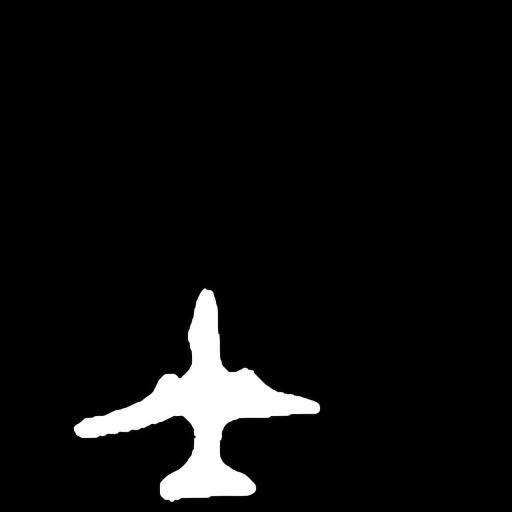} & \includegraphics[width=0.12\textwidth]{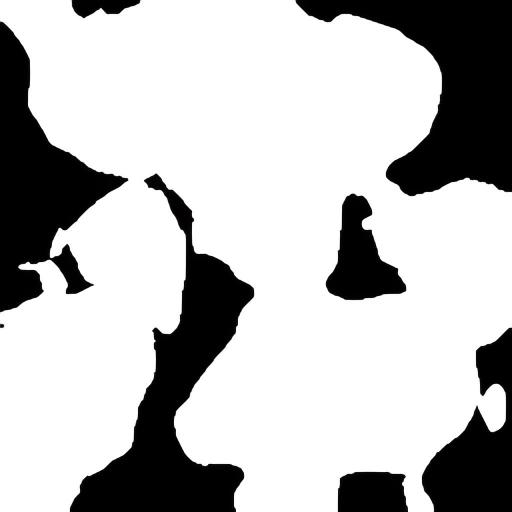} \\
    
            \raisebox{1.8\height}{\rotatebox[origin=c]{90}{Trufor}} &
            \includegraphics[width=0.12\textwidth]{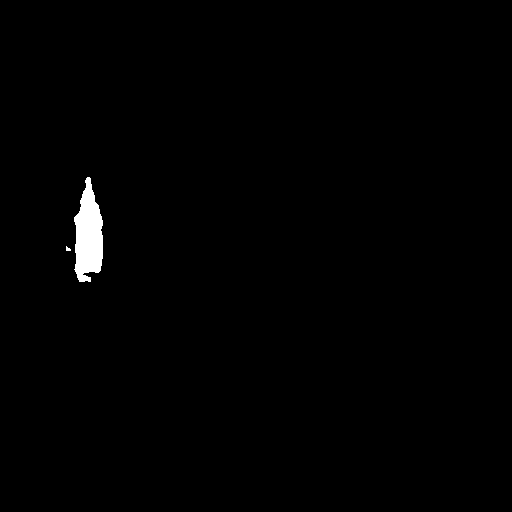} & \includegraphics[width=0.12\textwidth]{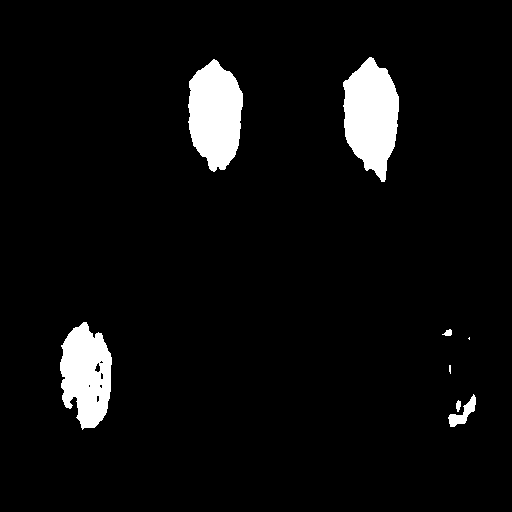} &
            \includegraphics[width=0.12\textwidth]{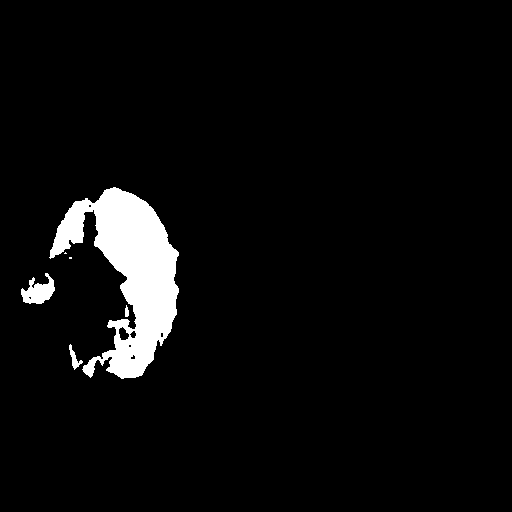} & \includegraphics[width=0.12\textwidth]{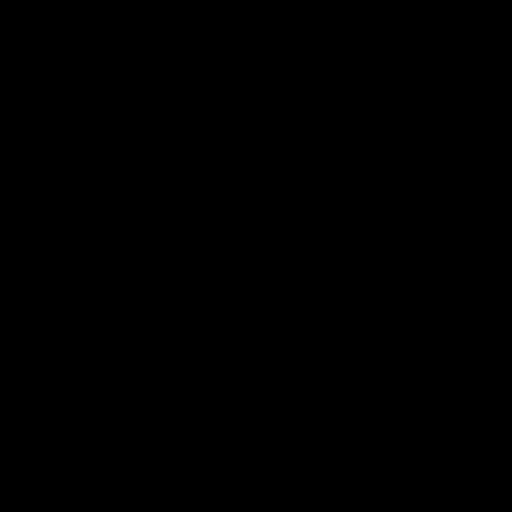} & \includegraphics[width=0.12\textwidth]{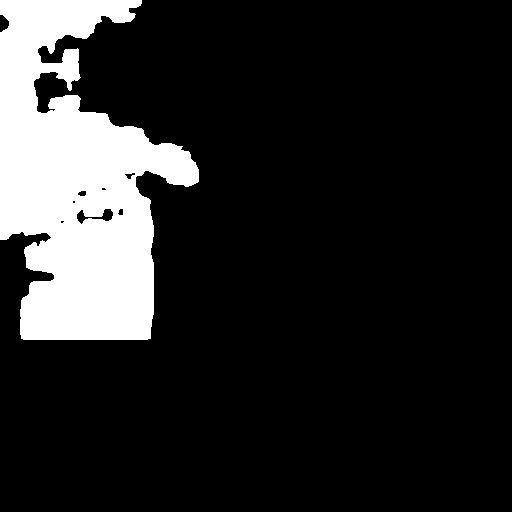} &
            \includegraphics[width=0.12\textwidth]{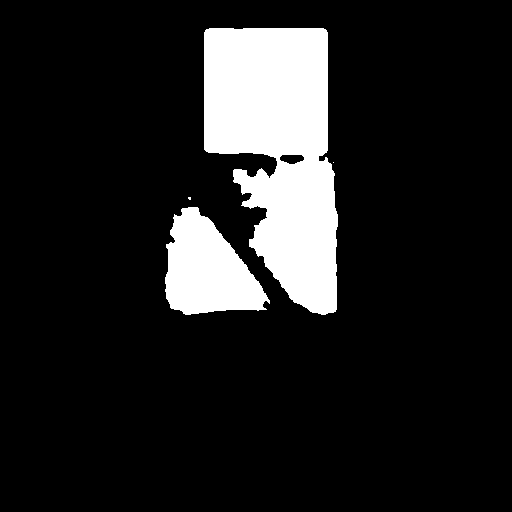} &
            \includegraphics[width=0.12\textwidth]{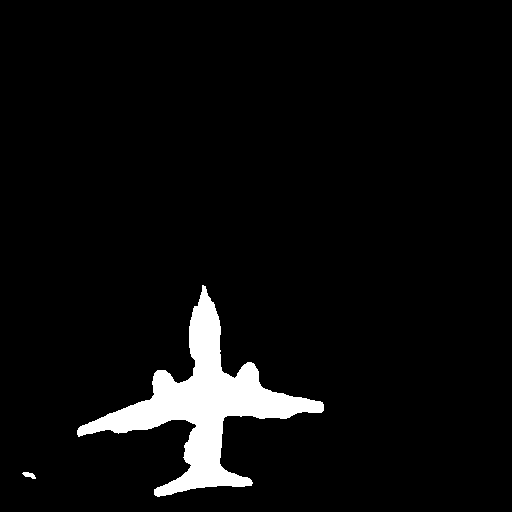} & \includegraphics[width=0.12\textwidth]{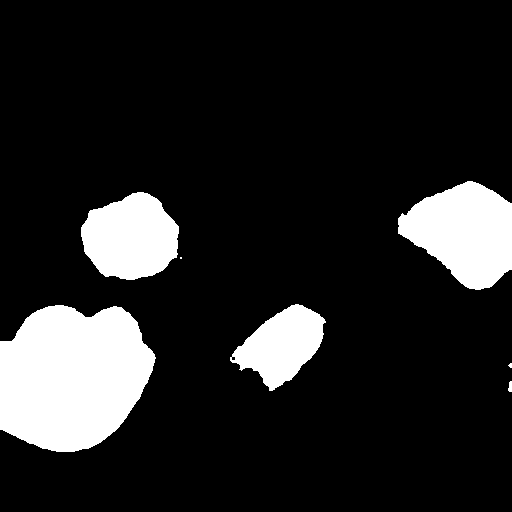} \\
    
            \raisebox{2\height}{\rotatebox[origin=c]{90}{MPC}} &
            \includegraphics[width=0.12\textwidth]{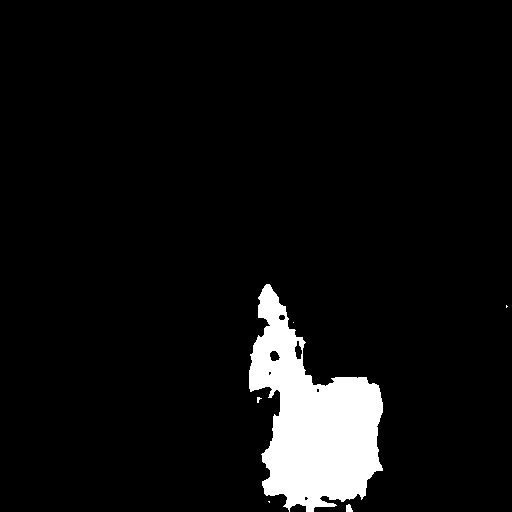} & \includegraphics[width=0.12\textwidth]{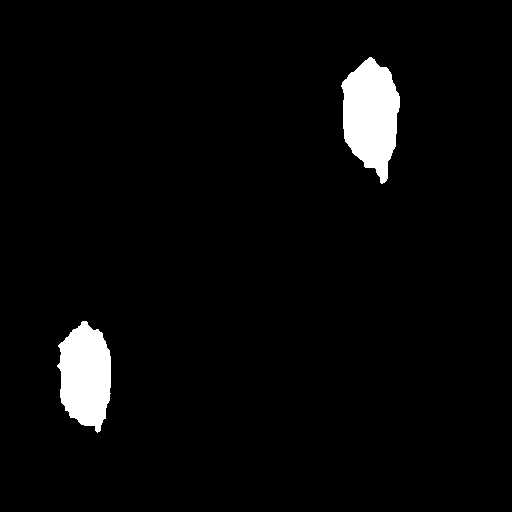} &
            \includegraphics[width=0.12\textwidth]{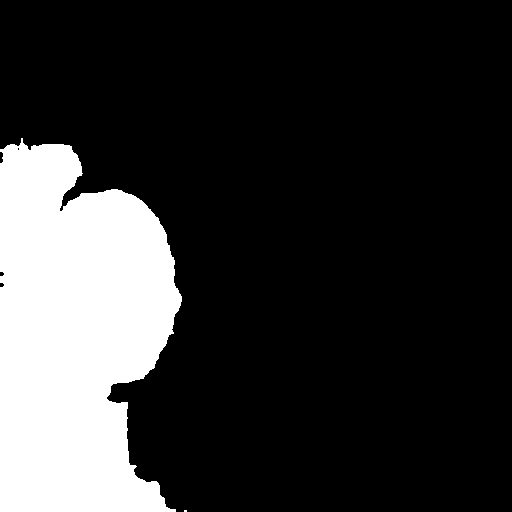} & \includegraphics[width=0.12\textwidth]{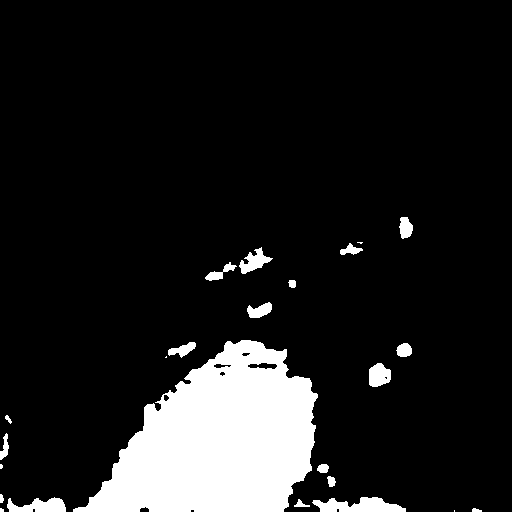} & \includegraphics[width=0.12\textwidth]{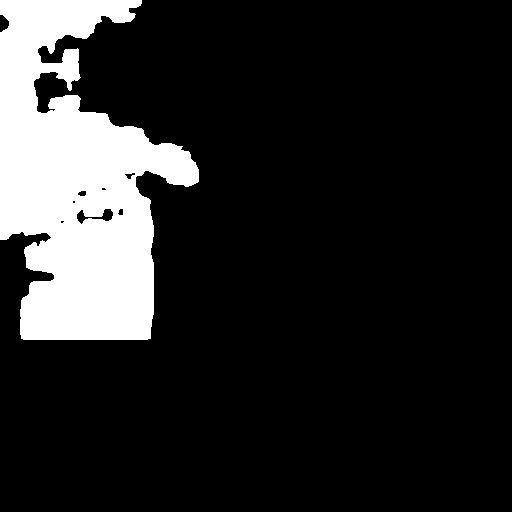} &
            \includegraphics[width=0.12\textwidth]{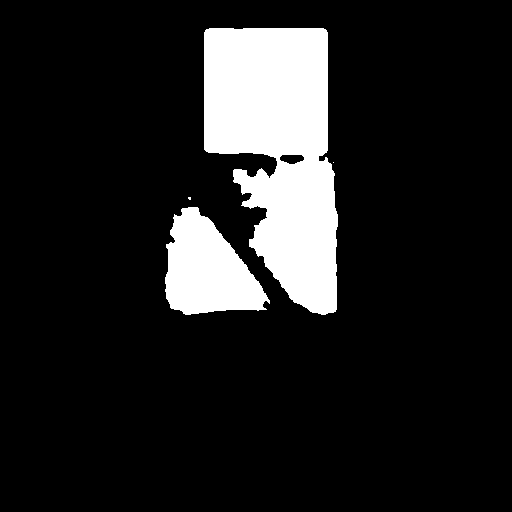} &
            \includegraphics[width=0.12\textwidth]{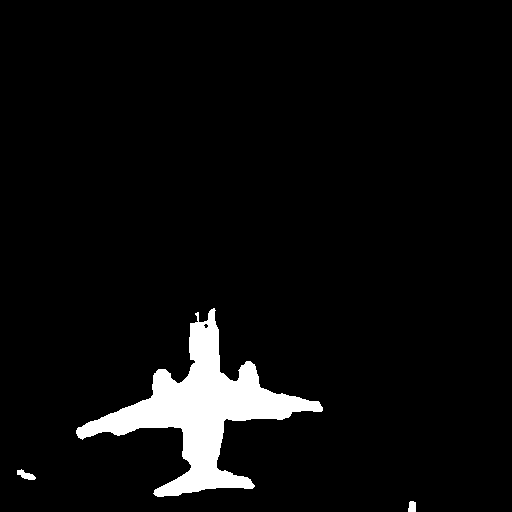} & \includegraphics[width=0.12\textwidth]{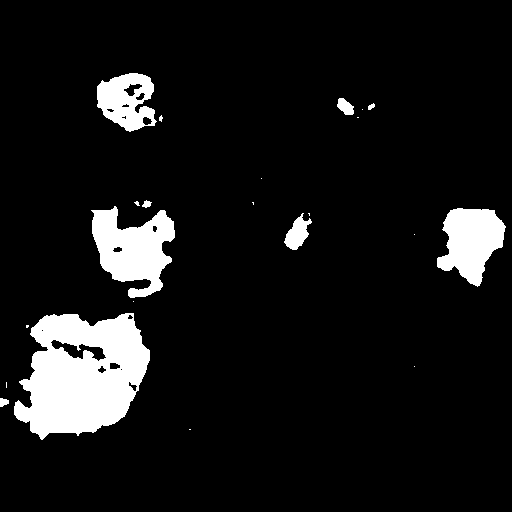}\\
    
            \raisebox{2\height}{\rotatebox[origin=c]{90}{Ours}} &
            \includegraphics[width=0.12\textwidth]{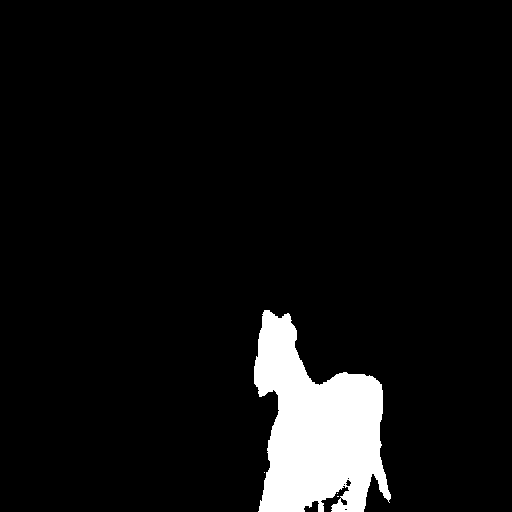} & \includegraphics[width=0.12\textwidth]{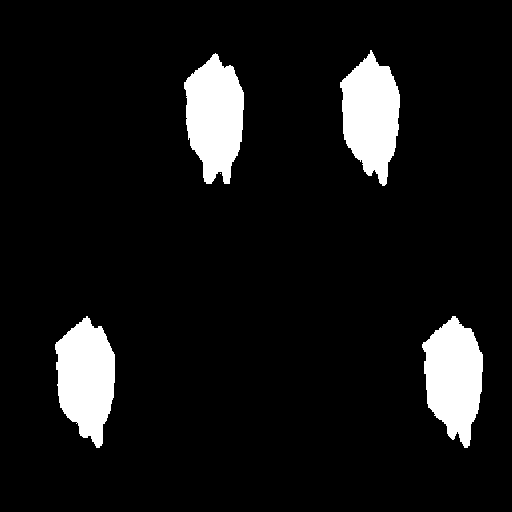} &
            \includegraphics[width=0.12\textwidth]{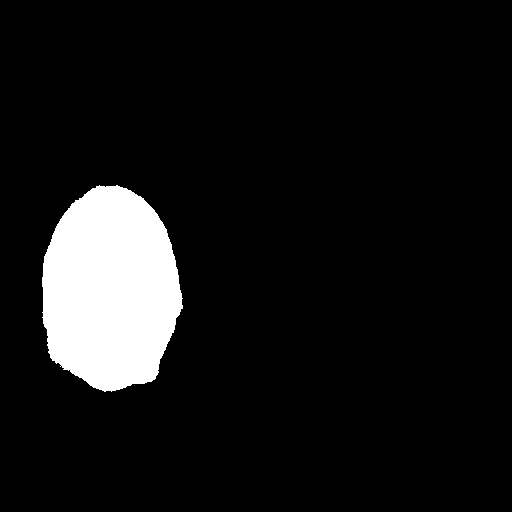} & \includegraphics[width=0.12\textwidth]{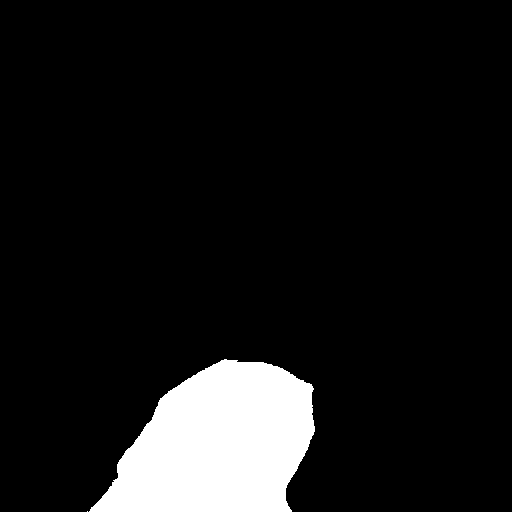} & \includegraphics[width=0.12\textwidth]{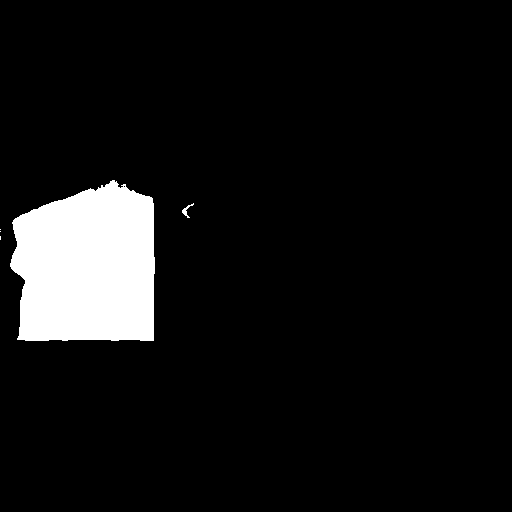} &
            \includegraphics[width=0.12\textwidth]{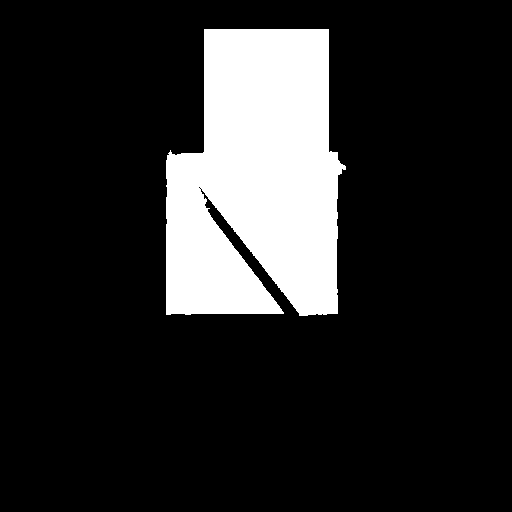} &
            \includegraphics[width=0.12\textwidth]{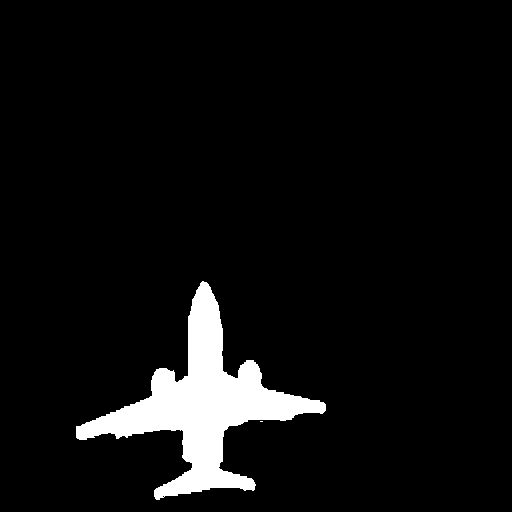} & \includegraphics[width=0.12\textwidth]{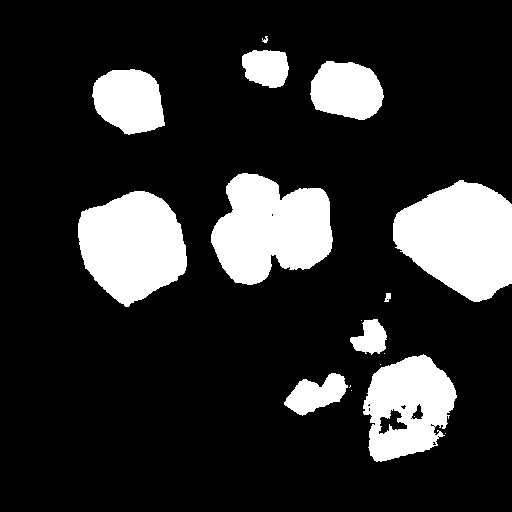}\\
        \end{tabular}
    }
    \caption{Qualitative comparison of forgery localization methods on example testing images. From left to right: two copy-move images, two splicing images, two removal images and two local AI-generated images. From top to bottom: tampered image, ground truth (GT), and the localization results from CAT-Net, TruFor, MPC and CBDiff.}
\end{figure}

\subsection{Comparison with State-of-the-Art Methods}
\begin{table}[!th]
    \centering
    \setlength{\tabcolsep}{2pt}
    \resizebox{0.47\textwidth}{!}{
        \begin{tabular}{l|c|cccccc|ccc}
            \toprule
            \multirow{2}{*}{Method-Year-Venue} & \multirow{2}{*}{Train-Num} & \multicolumn{2}{c}{CASIAv1}& 
            \multicolumn{2}{c}{Coverage}& \multicolumn{2}{c}{NIST16}& \multicolumn{2}{c}{Average}\\ \cline{3-10}
            & & F1& AUC&  F1&AUC&  F1&AUC& F1&AUC\\
            \midrule
            ELA-2007-HFS~\cite{ELA} & - & .214&.613& .222&.583& .236&.429& .222&.547\\
            NOI1-2009-IVC~\cite{NOI} & - &.263    &.612&.269     &.587&.285   &.487&.271    &.567\\
            CFA-2012-TIFS~\cite{CFA} & - &.207&.522&.190&.485&.174&.501&.195&.512\\
            RGB-N-2018-CVPR~\cite{RGB-N} & 42k &.408    &.795&.434     &.817&.718   &.937&.517&.846\\
            ManTraNet-2019-CVPR~\cite{ManTra-Net} & 42k       &.512&.817&.485&.819&.723&.962&.584&.867\\
            SPAN-2020-ECCV~\cite{SPAN} & 96k &.382    &.838&.558     &.937&.582&.961&.462&.887\\
            TransForensics-2021-ICCV~\cite{TransForensics} & 96k       &.479    &.850&.648     &.884&.753   &.984&.584&.899\\
            PSCC-Net-2022-TCSVT~\cite{PSCC-Net} & 100k &.554&.875&.723&.941&.742&.991&.630&.919\\
            MVSS-Net++-2022-TPAMI~\cite{MVSS-Net++} & 96k &.534    &.799&.764&.834&.849&.966&.657&.859\\
            SAT-2022-TCSVT~\cite{SAT} & 100k &.592&.843& \textbf{\textcolor{red}{.843}} & \textbf{\textcolor{red}{.985}} &.878&.990&.707&.902\\
            ObjectFormer-2022-CVPR~\cite{ObjectFormer} & 62k &.579    &.882&.758     &.957&.824&.996&.675&.926\\
            PCL-2023-TCSVT~\cite{PCL} & 100k &.467 &.751&.620&.917&.780&.946&.585&.829\\
            TBFormer-2023-SPL~\cite{TBFormer} & 150k &\underline{.696}&\textbf{\textcolor{red}{.955}} &.756&.896&.834&\underline{.997}&\underline{.748}&\underline{.966}\\
            HiFi-Net-2023-CVPR~\cite{HiFi-Net} & 100k &.616    &.885&\underline{.801}&.966&.850&.989&.709&.926\\
            UP-Net-2023-TCSVT~\cite{UP-Net} & 65k &.615    &.825&.577&.879&\underline{.916}&.995&.717&.887\\
            TANet-2023-TCSVT~\cite{TANet} & 60k &.614&.893&.782&.\underline{978}&.865&\underline{.997}&.711&.934\\
            DiffForensics-2024-CVPR~\cite{DiffForensics} & 50k       &586    &.886&.752&.915&.839&.942&.684&.907\\
            Ours-2025& 60k & \textbf{\textcolor{red}{.764}} & \underline{.954} &.756 &.976 & \textbf{\textcolor{red}{.939}} & \textbf{\textcolor{red}{.998}} & \textbf{\textcolor{red}{.824}} & \textbf{\textcolor{red}{.971}}\\
            \bottomrule
        \end{tabular}
    }
    \caption{Image Forgery Localization Performance F1 and AUC in Protocol 1. The best results are highlighted in red and the second-best results are underlined.}
    \label{tab:protocol1}
\end{table}

\begin{table*}[!th]
    \centering
    \resizebox{0.94\textwidth}{!}{
        \begin{tabular}{l|>{\centering\arraybackslash}p{0.07\linewidth}>{\centering\arraybackslash}p{0.07\linewidth}>{\centering\arraybackslash}p{0.07\linewidth}>{\centering\arraybackslash}p{0.07\linewidth}>{\centering\arraybackslash}p{0.07\linewidth}>{\centering\arraybackslash}p{0.07\linewidth}>{\centering\arraybackslash}p{0.07\linewidth}>{\centering\arraybackslash}p{0.07\linewidth}>{\centering\arraybackslash}p{0.08\linewidth}|>{\centering\arraybackslash}p{0.07\linewidth}}
            \toprule
            Method-Year-Venue & Params. & CASIAv1 & Coverage & Columbia & Grip & NIST16 & MISD & Wild & CocoGlide & Average \\
            \midrule
            \multicolumn{10}{c}{F1} \\
            \midrule
            CAT-Net-2022-IJCV~\cite{CAT-Net} & 114M & .710 & .285 & .792 & \textbf{\textcolor{red}{.446}} & .302 & .393 & .339 & .364 & .489 \\
            Trufor-2023-CVPR~\cite{Trufor} & 69M & .655 & .375 & \underline{.896} & .186 & .360 & .706 & \underline{.566} & \underline{.444} & .548 \\
            MPC-2025-TIFS~\cite{MPC} & \underline{45M} & .745 & \underline{.615} & \textbf{\textcolor{red}{.945}} & .222 & .436 & \underline{.721} & \textbf{\textcolor{red}{.591}} & .421 & \underline{.604} \\
            SAFIRE-2025-AAAI~\cite{Safire} & 97M & .327 & .576 & .864 & .134 & .410 & .615 & .451 & .458 & .440\\
            CBDiff-small-2025 & \textbf{\textcolor{red}{40M}} & \textbf{\textcolor{red}{.821}} & .551 & .817 & \underline{.377} & \textbf{\textcolor{red}{.510}} & .682 & .489 & .252 & .596 \\
            CBDiff-large-2025 & 140M & \underline{.799} & \textbf{\textcolor{red}{.681}} & .856 & .298 & \underline{.465} & \textbf{\textcolor{red}{.732}} & .485 & \textbf{\textcolor{red}{.468}} & \textbf{\textcolor{red}{.628}} \\
            \midrule
            \multicolumn{10}{c}{AUC} \\
            \midrule
            CAT-Net-2022-IJCV~\cite{CAT-Net} & 114M & .873 & .649 & .875 & .669 & .662 & .657 & .660 & .631 & .736 \\
            Trufor-2023-CVPR~\cite{Trufor} & 69M & .822 & .691 & .918 & .563 & .708 & .797 & .769 & .688 & .762 \\
            MPC-2025-TIFS~\cite{MPC} & \underline{45M} & .872 & \underline{.804} & \textbf{\textcolor{red}{.959}} & .596 & .745 & .818 & .776 & .674 & .793 \\
            SAFIRE-2025-AAAI~\cite{Safire} & 97M & .613 & .780 & .879 & .544 & .708 & .733 & .689 & .669 & .685 \\
            CBDiff-small-2025 & \textbf{\textcolor{red}{40M}} & \textbf{\textcolor{red}{.971}} & .802 & .897 & \textbf{\textcolor{red}{.781}} & \underline{.802} & \textbf{\textcolor{red}{.869}} & \textbf{\textcolor{red}{.868}} & \underline{.789} & \underline{.875} \\
            CBDiff-large-2025 & 140M & \underline{.962} & \textbf{\textcolor{red}{.951}} & \underline{.927} & \underline{.738} & \textbf{\textcolor{red}{.812}} & \underline{.859} & \underline{.821} & \textbf{\textcolor{red}{.851}} & \textbf{\textcolor{red}{.883}} \\
            \bottomrule
        \end{tabular}
    }
    \caption{Image Forgery Localization Performance F1 in Protocol 2. The best results are highlighted in red and the second-best results are underlined.}
    \label{tab:protocol2}
\end{table*}

We compare our method against 19 state-of-the-art image forgery localization approaches under both experimental protocols. Given the ability of our model to generate multiple plausible predictions, we produce eight predictions per test sample and aggregate them into a final result via a confidence-based voting mechanism. Under \textbf{Protocol 1}, Table~\ref{tab:protocol1} presents pixel-level localization performance in terms of F1-score and AUC across 16 existing methods. A dash ``--'' indicates methods that do not rely on training data, or lack publicly available information on training dataset size.

Overall, our method consistently achieves superior average performance across all test datasets, outperforming all competing methods in both F1-score and AUC. In particular, our method outperforms the second-best approach, TBFormer, by 7.4\% in average F1-score. It achieves top results on CASIAv1 and NIST16, with margins of 6.8\% over TBFormer and 2.3\% over UP-Net, respectively. Strong performance is also observed on the Coverage dataset. These results demonstrate the strong cross-dataset generalization capability and robustness of our approach. Under \textbf{Protocol 2}, Table~\ref{tab:protocol2} shows F1 and AUC scores for four methods trained on comparable-sized datasets. Our method significantly outperforms all competitors on CASIAv1, Coverage, NIST16, MISD, and CocoGlide. On average, we outperform the best prior method by 0.024 in both F1-score and AUC. In summary, the results clearly indicate that our method outperforms current state-of-the-art approaches in both performance and generalization on the image forgery localization task.

\subsection{Robustness Evaluation}

\begin{table}
    \centering
    \resizebox{0.47\textwidth}{!}{
        \begin{tabular}{l|c|ccc|c}
            \toprule
            Method & OSNs & CASIAv1 & Columbia & NIST16 & Average \\
            \midrule
            CAT-Net~\cite{CAT-Net} & \multirow{4}{*}{Facebook} & .633 & .918 & .151 & .499 \\
            Trufor~\cite{Trufor} &  & .672 & .749 & .353 & .573 \\
            MPC~\cite{MPC} &  & .709 & \textbf{\textcolor{red}{.958}} & .429 & .639 \\
            Ours &  & \textbf{\textcolor{red}{.767}} & .831 & \textbf{\textcolor{red}{.449}} & \textbf{\textcolor{red}{.667}}\\
            \hline
            CAT-Net~\cite{CAT-Net} & \multirow{4}{*}{Wechat} & .139 & .848 & .191 & .224\\
            Trufor~\cite{Trufor} &  & .569 & .773 & .351 & .515 \\
            MPC~\cite{MPC} &  & .612 & \textbf{\textcolor{red}{.947}} & .423 & .581 \\
            Ours &  & \textbf{\textcolor{red}{.683}} & .833 & \textbf{\textcolor{red}{.430}} & \textbf{\textcolor{red}{.613}}\\
            \hline
            CAT-Net~\cite{CAT-Net} & \multirow{4}{*}{Weibo} & .425 & .921 & .208 & .400 \\
            Trufor~\cite{Trufor} &  & .637 & .800 & .332 & .550 \\
            MPC~\cite{MPC} &  & .707 & \textbf{\textcolor{red}{.948}} & \textbf{\textcolor{red}{.433}} & .638 \\
            Ours &  & \textbf{\textcolor{red}{.751}} & .830 & .419 & \textbf{\textcolor{red}{.647}}\\
            \hline
            CAT-Net~\cite{CAT-Net} & \multirow{4}{*}{Whatsapp} & .423 & .921 & .201 & .396 \\
            Trufor~\cite{Trufor} &  & .663 & .747 & .323 & .557 \\
            MPC~\cite{MPC} &  & .696 & \textbf{\textcolor{red}{.950}} & \textbf{\textcolor{red}{.438}} & .634 \\
            Ours &  & \textbf{\textcolor{red}{.765}} & .824 & .430 & \textbf{\textcolor{red}{.659}}\\
            \bottomrule
        \end{tabular}
    }
    \caption{Robustness performance F1 on  Social Network Post-processing. The best results are highlighted in red.}
    \label{tab:robustness}
\end{table}

To assess robustness against post-processing artifacts commonly introduced by Online Social Networks (OSNs), we conduct a systematic evaluation following the setup in~\cite{wu2022robust}. Specifically, we evaluate forged images that have undergone compression and transformation by four major OSNs: Facebook, Weibo, WeChat, and WhatsApp. The test is performed using three widely adopted datasets: CASIAv1, Columbia, and NIST16. Table~\ref{tab:robustness} presents the F1-scores and AUC metrics obtained by four representative methods on these OSN-degraded images. Our method consistently achieves the highest performance across all test cases, with especially notable improvements on CASIAv1 and NIST16. The robustness of our approach stems from its diffusion-based architecture, which generates localization maps through a progressive, multi-step generation process. This design helps suppress the distortion caused by OSN post-processing and enhances the detection of subtle forgery traces. Even without considering the benefit of producing diverse outputs, our method proves to be highly effective in mitigating the real-world challenge of forgery propagation across social platforms.

\subsection{Ablation Studies}

\begin{table}
    \centering
    \setlength{\tabcolsep}{2pt}
    \resizebox{0.47\textwidth}{!}{
        \begin{tabular}{l|c|ccc|c}
            \toprule
            & Variant & CAS. & COV. & NIS. & Ave. \\
            \midrule
            \multirow{2}{*}{Noise} & Gaussian & .759 & .431 & .396 & .608 \\
            & Bernoulli & \textbf{\textcolor{red}{.821}} & \textbf{\textcolor{red}{.567}} & \textbf{\textcolor{red}{.514}} & \textbf{\textcolor{red}{.696}}\\
            \hline
            \multirow{2}{*}{Noise Schedule} & Linear & .765 & .529 & .476 & .647 \\
            & Cosine & \textbf{\textcolor{red}{.821}} & \textbf{\textcolor{red}{.567}} & \textbf{\textcolor{red}{.514}} & \textbf{\textcolor{red}{.696}}\\
            \hline
            \multirow{4}{*}{Time Steps} & 100 & .763 & .511 & .467 & .642 \\
            & 50 & \textbf{\textcolor{red}{.821}} & \textbf{\textcolor{red}{.567}} & \textbf{\textcolor{red}{.514}} & \textbf{\textcolor{red}{.696}}\\
            & 25 & .799 & .535 & .480 & .669 \\
            & 10 & .799 & .441 & .457 & .654 \\
            \hline
            \multirow{7}{*}{Conditional Strategy} & w/o TSCAttention & .693 & .364 & .363 & .554 \\
            & w/o DINO \& TSCAttention & .682 & .353 & .353 & .544 \\
            & w/o Image & .801 & .467 & .466 & .661 \\
            & w/o Image \& TSCAttention & .373 & .253 & .216 & .309 \\
            & w/o Noiseprint++ & .811 & .524 & .479 & .675 \\
            & All & \textbf{\textcolor{red}{.821}} & \textbf{\textcolor{red}{.567}} & \textbf{\textcolor{red}{.514}} & \textbf{\textcolor{red}{.696}}\\
            \hline
            \multirow{3}{*}{Attention} & w/o Attention & .693 & .364 & .363 & .554 \\
            & w CrossAttention & .787 & .542 & .456 & .654 \\
            & w TSCAttention & \textbf{\textcolor{red}{.821}} & \textbf{\textcolor{red}{.567}} & \textbf{\textcolor{red}{.514}} & \textbf{\textcolor{red}{.696}}\\
            \bottomrule
        \end{tabular}
    }
    \caption{Comparison of localization performance for ablation studies on a lightweight CBDiff model with 45M parameters. The best results are highlighted in red.}
    \label{tab:ablation}
\end{table}

To validate the effectiveness of the proposed CBDiff framework, we conducted a series of systematic ablation studies on CBDiff-small, where the input image resolution is set to $256\times256$ and the total number of parameters is approximately 40M. All results are reported in Table~\ref{tab:ablation}, enabling an in-depth understanding of each component's contribution.

\noindent\textbf{Effectiveness of Bernoulli Noise.} We hypothesize that since forgery masks are inherently binary, using Gaussian noise as in conventional diffusion models unnecessarily enlarges the target space, making learning more difficult. We compare models initialized with Bernoulli and Gaussian noise. As shown in Table~\ref{tab:ablation}, Bernoulli noise significantly outperforms Gaussian noise across all datasets, demonstrating its superior compatibility with binary prediction tasks and the overall effectiveness of CBDiff.

\noindent\textbf{Influence of Noise Schedule.} We compare two noise schedules for Bernoulli noise: a linear schedule ($\text{start}=0.01$, $\text{end}=0.2$) and a cosine schedule ($s=0.008$). The cosine schedule provides better performance, as indicated in Table~\ref{tab:ablation}, suggesting a more effective denoising process.

\noindent\textbf{Influence of Time Steps.} To balance generation quality and training efficiency, we evaluate $T=\{100, 50, 25, 10\}$ time steps. Results show that 50 steps offer the best trade-off, as detailed in Table~\ref{tab:ablation}, aligning well with the moderate complexity of the localization task.

\noindent\textbf{Influence of Conditional Strategy.} CBDiff uses three types of conditioning: the forged image, DINO-extracted multi-scale features, and Noiseprint++ forensic features. Ablation results in Table~\ref{tab:ablation} show performance drops when any of these are removed. Notably, removing DINO features or the TSCAttention module causes significant degradation, and combining DINO without TSCAttention offers only marginal gains—highlighting the importance of both. Removing image-level input or Noiseprint++ also harms performance, although to a lesser extent. 

\noindent\textbf{Influence of Attention Mechanism.} The experimental results in Table~\ref{tab:ablation} demonstrate that replacing basic CrossAttention with TSCAttention further confirms its effectiveness in temporal modeling.

\subsection{Limitations}
CBDiff-large contains 140M parameters, making it significantly larger than prior methods like MPC and TruFor. Its iterative denoising process also incurs higher inference time. To address this, we developed CBDiff-small, a 40M-parameter variant that achieves comparable performance to MPC. Additionally, a one-step inference mode can be used to reduce latency, though it may lead to reduced output diversity.
\section{Conclusion}
In this paper, we introduce CBDiff, a novel conditional Bernoulli diffusion model designed for image forgery localization. Unlike traditional discriminative models that produce a single deterministic mask, CBDiff generates multiple diverse and plausible predictions. This capability effectively addresses the inherent uncertainty and variability of manipulated regions. By substituting Gaussian noise with Bernoulli noise and incorporating the TSCAttention module, our approach is better suited to the binary nature of forgery masks while also leveraging rich semantic and temporal cues. Through extensive experiments across various benchmarks, we demonstrate that CBDiff not only achieves state-of-the-art performance but also exhibits robust generalization, superior resistance to post-processing artifacts, and high interpretability. This work introduces a new generative paradigm for forensic analysis and opens up new avenues for research in uncertainty-aware visual forensics. While CBDiff marks a significant advancement in uncertainty-aware forgery localization, future work could explore integrating multimodal forensic cues, such as metadata or textual descriptions, to improve context-aware manipulation detection. Additionally, developing lightweight and real-time versions of CBDiff for edge devices and mobile applications could enhance deployment.

{
    \small
    \bibliographystyle{ieeenat_fullname}
    \bibliography{main}
}


\end{document}